\documentclass[letterpaper]{article} 
\usepackage{aaai25}  
\usepackage{times}  
\usepackage{helvet}  
\usepackage{courier}  
\usepackage[hyphens]{url}  
\usepackage{graphicx} 
\usepackage{natbib}  
\usepackage{caption} 
\usepackage{algorithm}
\usepackage{algorithmic}
\usepackage{amsmath}
\usepackage{booktabs}
\usepackage{subcaption}
\usepackage{siunitx}
\usepackage{comment}

\usepackage{newfloat}
\usepackage{listings}
\DeclareCaptionStyle{ruled}{labelfont=normalfont,labelsep=colon,strut=off} 
\floatstyle{ruled}
\newfloat{listing}{tb}{lst}{}
\floatname{listing}{Listing}
\title{Generic Guard AI in Stealth Game with Composite Potential Fields}
\author{
    Kaijie Xu, Clark Verbrugge\\
}
\affiliations{
    Department of Computer Science, McGill University\\
    Montreal, Quebec, Canada\\
    kaijie.xu2@mail.mcgill.ca, clump@cs.mcgill.ca
}

\usepackage{bibentry}
\nocopyright

\begin{document}

\maketitle

\begin{abstract}
Guard patrol behavior is central to the immersion and strategic depth of stealth games, while most existing systems rely on hand-crafted routes or specialized logic that struggle to balance coverage efficiency and responsive pursuit with believable naturalness. We propose a generic, fully explainable, training-free framework that integrates global knowledge and local information via Composite Potential Fields, combining three interpretable maps—Information, Confidence, and Connectivity—into a single kernel-filtered decision criterion. Our parametric, designer‐driven approach requires only a handful of decay and weight parameters—no retraining—to smoothly adapt across both occupancy‐grid and NavMesh‐partition abstractions. We evaluate on five representative game maps, two player-control policies, and five guard modes, confirming that our method outperforms classical baseline methods in both capture efficiency and patrol naturalness. Finally, we show how common stealth mechanics—distractions and environmental elements—integrate naturally into our framework as sub modules, enabling rapid prototyping of rich, dynamic, and responsive guard behaviors.
\end{abstract}

%

\section{Introduction}

Stealth games center on the core interactions where players aim to remain undetected by or strategically overcome patrolling guards. Guard patrol behavior is therefore important to the stealth experience, directly influencing perceived challenge, strategic depth, and narrative immersion \cite{isla2005handling,al2020dynamic}. In current games, these patrols are usually costly hand‐scripted and highly repetitive with low maintainability—enemies follow fixed routes or simple state‐machine routines that are easily memorized and hard to adapt to new environments. This over predictability undermines gameplay: when guards consistently fall for the same distractions or never vary their routes, expert players feel they are exploiting a script rather than outsmarting a dynamic adversary. Previous analysis confirms that in many stealth titles, AI agents stick to predictable patterns with minimal response to player actions \cite{isla2013third}. The result is a loss of immersion and reduced player engagement as the illusion of intelligence breaks down.

Prior attempts to improve guard patrols span various approaches, including probabilistic models to track player positions~\cite{isla2006probabilistic, isla2013third, hladky2008evaluation, 9619054}, dynamic coverage techniques that avoid repetitive paths~\cite{xu2014generative, al2020dynamic}, and reinforcement learning for adaptive multi-agent behaviors~\cite{chia2022artist}. While these methods address predictability, responsiveness, or coverage, each has trade-offs: probabilistic approaches can be computationally heavy or overly specialized, dynamic coverage sacrifices pursuit efficiency, and learned methods lack interpretability, complicating design. Hence, there remains a clear need for a unified, explainable, and designer-friendly solution that effectively balances exploration, pursuit responsiveness, and natural guard behavior.

To mitigate these limitations, we propose a generic, explainable, and training-free guard AI framework based on \emph{composite potential fields}. This framework integrates three interpretable maps—\emph{information}, \emph{confidence}, and \emph{connectivity}—into a combined decision-making process controlled by simple, designer-tunable parameters. By adjusting these parameters, designers can effortlessly achieve diverse, adaptive guard behaviors without extensive retraining across both occupancy‐grid and NavMesh‐partition abstractions.

We evaluate our approach across multiple representative game environments, player behaviors, and guard strategies, confirming significant improvements over classical baselines in both patrol efficiency and realism. Our results indicate that our framework achieves strong performance and enables easy incorporation of common stealth mechanics such as distractions, decoys, and environmental interactions, promoting richer, more dynamic gameplay. This work not only addresses immediate challenges in stealth game AI but also offers a basis for future innovations in believable and adaptable common virtual agents. Our key contributions are:
\begin{itemize}
  \item A generic, explainable guard AI based on composite potential fields integrating global and local information.
  \item A kernel-filtered decision process enabling effective balance between exploration and pursuit without training.
  \item Detailed experimental analysis showing better performance in diverse scenarios.
  \item Straightforward extensions to incorporate common stealth game mechanics in the same unified framework.
\end{itemize}

\section{Background}

\subsection{Guard Behavior in Games}

Guard behavior shapes realism, difficulty, and immersion in stealth games~\cite{al2023evaluating,al2023investigating}. Traditional FSMs and fixed patrol routes are easy to implement but produce predictable and repetitive patterns, enabling players to quickly exploit their limitations. This predictability reduces the challenge and breaks the illusion of intelligent adversaries, weakening long-term engagement~\cite{isla2005handling,isla2013third}.

Dynamic, adaptive approaches have been proposed to overcome these shortcomings. Isla’s~(\citeyear{isla2006probabilistic}, \citeyear{isla2013third}) occupancy maps guide search by maintaining a probability distribution over the player’s likely positions, and Hladký and Bulitko~(\citeyear{hladky2008evaluation}) employ probabilistic inference to anticipate future player movements. Environmental skeletons support coordinated multi-agent search when the player is hidden~\cite{9619054}. Al Enezi and Verbrugge’s~(\citeyear{al2020dynamic}, \citeyear{al2023evaluating})  ``staleness'' metric—where each region’s score grows with time since its last visit—drives guards toward under-patrolled areas, balancing control and exploration. We adopt this staleness model as a strong, interpretable baseline in our study. Other methods include generative optimization for guard placement~\cite{xu2014generative}, behavior trees for context-aware transitions, and AI personalities for varied NPC traits~\cite{cernak2021tower}. Deep reinforcement learning coordinates complex ambushes \cite{chia2022artist}, but its policies are opaque, computationally expensive, and hard to retune.

Thus, the challenge remains to develop guard AI systems that balance adaptiveness, coverage efficiency, and natural responsiveness, without becoming overly predictable, opaque, or resource-intensive. Our work directly addresses this gap by proposing a generic, explainable, training-free framework using composite potential fields.

\subsection{Potential Fields in AI}
Artificial Potential Fields (APFs), originally introduced in robotics~\cite{khatib1986real},  provide a clear, real-time way to steer agents through complex spaces. In PFs, agents are represented as particles influenced by attractive forces from goals and repulsive forces from obstacles. The combination of these forces produces a vector field guiding agents smoothly toward objectives while avoiding collisions~\cite{samodro2023artificial}. PFs have been extensively used in robotics for tasks like obstacle avoidance, path planning, and group coordination. However, traditional PFs can encounter issues such as local minima, where conflicting forces cause agents to become temporarily trapped between multiple influences. Solutions include heuristic methods, stochastic perturbations, and memory-based techniques~\cite{alarabi2024multi}.

Game AI has adopted the same idea for varied tasks. Beyond fundamental NPC pathfinding and recreating pedestrian social forces~\cite{helbing1995social}, these field-based techniques have been used in Real-Time Strategy games for coordinating group movements~\cite{hagelback2008using, danielsiek2008intelligent} and enabling micro control such as kiting~\cite{uriarte2012kiting}. Influence maps have also been combined with machine learning to analyze game states in StarCraft~\cite{SANCHEZRUIZ201729}. More recently, PFs are utilized through multi-agent reinforcement learning for generative subgoal orientation~\cite{li2024generative}. The clear potential of PFs to inform agent behavior across various game scenarios motivates our approach. We focus these concepts by developing \emph{composite potential fields} designed for the complex decision-making of guards in stealth games.

\section{Methods}
In this section, we introduce our level abstractions, the composite potential fields, kernel-filtered decision process, and adaptive weights that together govern guard navigation.

\subsection{Environment Abstractions}
\paragraph{Occupancy Grid}
We discretize the level into an $r\times c$ array of square cells $c_{ij}$. Both player and guards occupy exactly one cell at a time. At each decision tick, an agent may move to one of its four orthogonal neighbors (up, down, left, right), provided that the cell is walkable. The agent’s new position is snapped to the center of the target cell, and its grid coordinates are updated accordingly.

\paragraph{Partition Nodes}
We build a graph $G=(V,E)$ from Unity’s NavMesh triangulation: each triangle’s centroid is a node $v\in V$, and edges connect nodes sharing a mesh edge. Guards and player are represented by NavMeshAgent components. When selecting a target node, the agent uses Unity’s pathfinder to move continuously along the mesh until the centroid is reached. This leads to smooth, obstacle‐aware pathfinding over the partition graph.

\subsection{Composite Potential Fields}

\begin{figure}[!t] 
  \centering
  \subfloat[Base Maze Layout with Agents]{
    \includegraphics[width=0.45\linewidth]{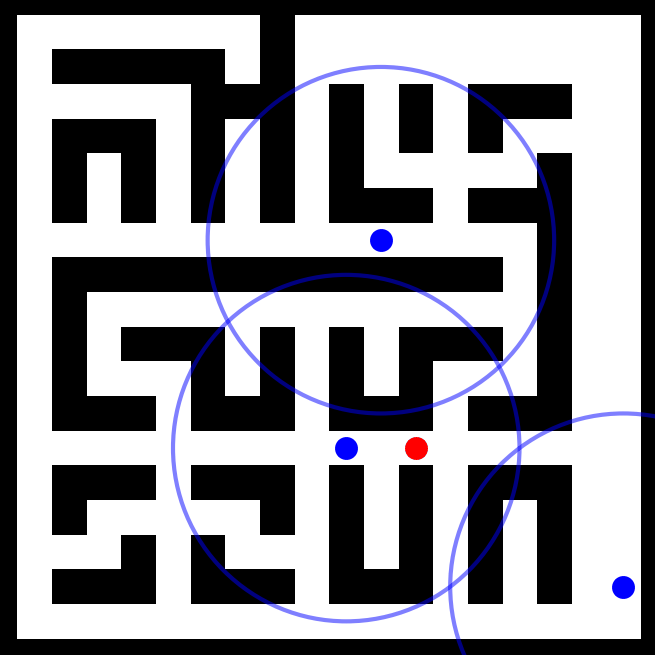}%
    \label{fig:viz_base_maze_agents}
  }
  \hfill 
  \subfloat[Information Field $I(n)$ Example]{
    \includegraphics[width=0.45\linewidth]{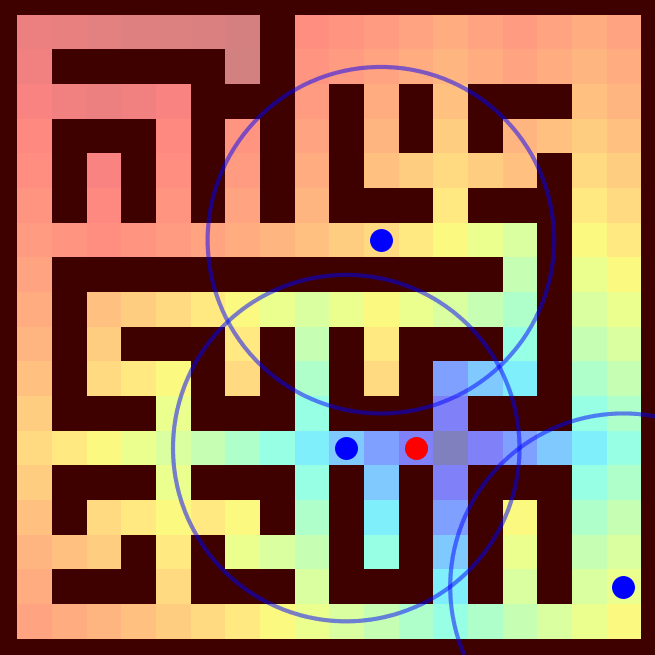}%
    \label{fig:viz_info_field_example}
  }
  \vspace{1ex} 
  \subfloat[Confidence Field $C(n)$ Example]{
    \includegraphics[width=0.45\linewidth]{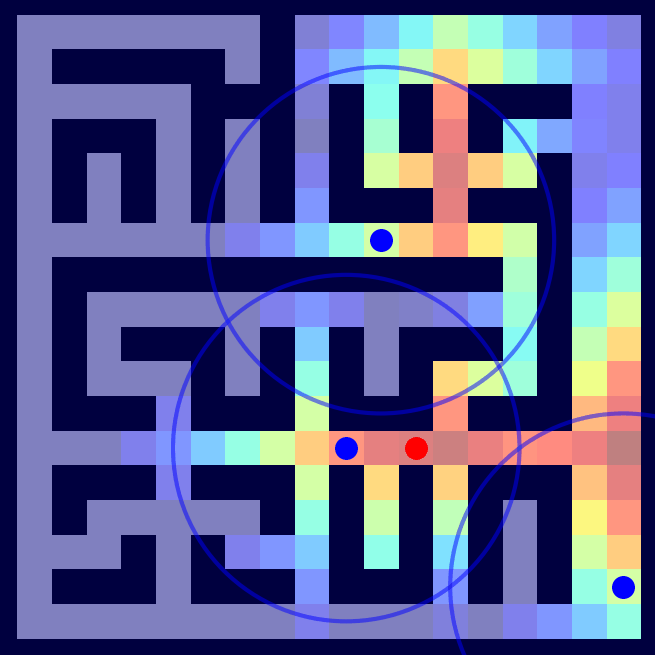}%
    \label{fig:viz_conf_field_example}
  }
  \hfill 
  \subfloat[Connectivity Field $N(n)$ Example]{
    \includegraphics[width=0.45\linewidth]{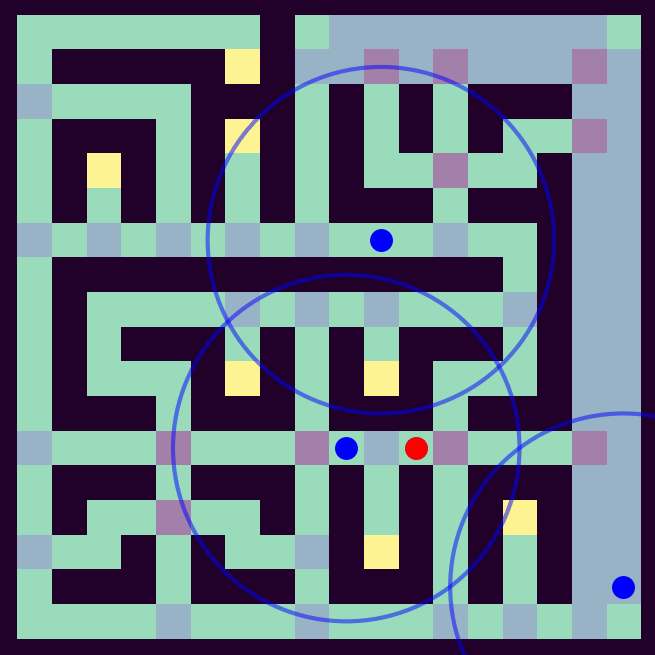}%
    \label{fig:viz_connect_field_example}
  }
  \caption{Visualization of the individual components of the Composite Potential Field within our test maze. (a) The base maze layout showing the player (red dot) and three guards (blue dots with vision/influence radii). (b) The Information Field, emanating from the player's current (or last known) position (red dot), creates a strong attractive potential (e.g., represented by cooler colors like blue) that diminishes with distance. (c) The Confidence Field, showing areas influenced by guard presence. Higher confidence (e.g., warmer colors) indicates recently patrolled zones, making them less immediately attractive for revisits by other guards, thus encouraging broader coverage. (d) The static Connectivity Field, where structurally significant areas are highlighted. For instance, more open or connected cells might be represented by one range of values (e.g., lighter green), while bottlenecks or dead-ends (yellowish patches in this example) are differentiated, influencing patrol routes based on their assigned weight.}
  \label{fig:composite_fields_viz}
\end{figure}

Our framework combines three distinct but complementary potential fields—Information, Confidence, and Connectivity—to guide guard behavior. Figure~\ref{fig:composite_fields_viz} provides a snapshot of these individual fields operating within a handcrafted test maze, illustrating their respective contributions to the overall decision-making landscape. The player is represented by a red dot, and guards by blue dots. The subsequent paragraphs detail the formulation and role of each field.

\paragraph{Information Field \(I(n)\)}
The Information Field, \(I(n)\), represents the ``scent'' of the player's presence, guiding guards toward potential areas of interest. At each update step, existing \(I(n)\) values are first temporally decayed:
\[
I(n) \;\leftarrow\; \rho\,I(n),
\]
where \(\rho\in(0,1)\) is the temporal decay rate, shared with the Confidence Field, causing the influence of old information to gradually wane over time if not refreshed. The primary update to this field occurs whenever any guard first detects the player at node \(p_t\) (the red dot in Figure~\ref{fig:viz_base_maze_agents}). At this point, \(I(n)\) is reset to 0 for \emph{all} nodes, and a negative-valued potential, indicating strong attraction, is propagated from \(p_t\) via Breadth-First Search (BFS). This newly propagated potential, visualized in Figure~\ref{fig:viz_info_field_example}, spatially decays with distance:
\[
I(n) \;=\; -I_0\,\gamma_I^{\,d(p_t,n)},\quad \forall\,n\in V, 
\]
where \(I_0>0\) is the initial magnitude of attraction, \(\gamma_I\in(0,1)\) is the spatial decay factor, and \(d(p_t,n)\) is the graph or grid distance from \(p_t\) to node \(n\). This event-driven update, combined with continuous temporal decay, ensures that guards are strongly guided towards fresh intelligence, while outdated clues diminish in relevance.

\paragraph{Confidence Field \(C(n)\)}
The Confidence Field represents how well areas are currently or have recently been covered by guards. As with the Information Field, existing \(C(n)\) values are temporally decayed at each update using the rate \(\rho\). Subsequently, for each guard currently at position \(g_t\) (blue dots in Figure~\ref{fig:viz_base_maze_agents}), a positive potential (representing an injection of confidence or ``patrolled'' status) is added to all its neighbor nodes, decaying with distance from the guard:
\[
C(n)\;\leftarrow \rho\,C(n)\;+\;\sum_{g\in \text{Guards}}\Bigl(C_0\,\gamma_C^{\,d(g_t,n)}\Bigr),
\]
where \(C_0>0\) is the per-guard injection magnitude—i.e., the confidence boost a single guard deposits at its node—and \(\gamma_C\in(0,1)\) is the spatial distance-decay factor for confidence. Figure~\ref{fig:viz_conf_field_example} illustrates how guard presence (blue dots) creates regions of higher confidence (warmer colors), effectively discouraging immediate revisits to already covered areas and promoting patrol diversity. Thus, each node’s confidence reflects a combination of its prior (decayed) coverage and current guard influences.

\paragraph{Connectivity Field \(N(n)\)}
This field is computed once from the static map topology to quantify the local structural properties of each node \(n\), as depicted in Figure~\ref{fig:viz_connect_field_example}:
\[
N(n)\;=\;\kappa\;-\;|\mathrm{Nbr}(n)|,
\]
where \(\mathrm{Nbr}(n)\) is the set of adjacent (neighboring) nodes to \(n\), and \(\kappa\) is a constant, typically set to the maximum degree observed in the graph (or a fixed value like 4 for grid environments). Nodes with fewer neighbors, such as those in dead-ends or narrow corridors (potentially shown as distinct patches, e.g., yellowish areas in Figure~\ref{fig:viz_connect_field_example}), thus receive higher \(N(n)\) values. This field allows guards to prioritize or deprioritize such structural features based on the sign and magnitude of their weights in the composite potential, influencing path choices related to exploration or strategic positioning.

\paragraph{Composite Potential \(P(n)\)}
At each decision point, a guard computes the \emph{Composite Potential} \(P(n)\) for all its candidate next nodes. This scalar value is a weighted linear combination of the three individual fields:
\[
P(n)\;=\;w_I(t)\,I(n)\;+\;w_C(t)\,C(n)\;+\;w_N(t)\,N(n),
\]
where \(w_I(t)\), \(w_C(t)\), and \(w_N(t)\) are the dynamically adjusted weights for the Information, Confidence, and Connectivity fields, respectively. This combined potential \(P(n)\) then serves as the input to the kernel-filtered decision process, which ultimately determines the guard's next movement target. The dynamic interplay of these fields and their weights allows the AI to fluidly transition between behaviors such as aggressive pursuit, systematic area patrolling, and strategic exploration of key topological features.

\subsection{Kernel-Filtered Graph Decision}
\label{sec:kernel}

\paragraph{Candidate Set Selection}
Before evaluating potential, each guard at current node \(n_t\) assembles a local candidate set \(\mathcal{C}(n_t)\):
\begin{itemize}
  \item \textbf{Grid Environment}  
    \(\mathcal{C}(n_t)\) comprises the four orthogonal neighbor cells, ensuring one‐step moves.
  \item \textbf{Partition Environment}: To enable smoother, multi-hop planning, \(\mathcal{C}(n_t)\) is primarily constructed by performing a BFS from the guard's current node \(n_t\). This search expands until the accumulated path length to potential candidates reaches a predefined lookahead threshold \(R_{\text{edge}}\) (e.g., typically set to 2-3 times the average inter-node distance, tuned empirically to balance foresight and computational cost). The nodes reached at this threshold form the primary candidates. If this BFS-based search yields no viable candidates within the \(R_{\text{edge}}\) limit (e.g., in very confined spaces or near the global boundary of the explorable area), the candidate set \(\mathcal{C}(n_t)\) defaults to include only the directly adjacent (neighboring) nodes to \(n_t\).

\end{itemize}
This constrained set focuses computation on reachable, locally meaningful waypoints.

\paragraph{Kernel-Filtered Evaluation and Target Selection}
Each candidate \(x\in\mathcal{C}(n_t)\) is scored by a kernel‐smoothed potential
\[
K(x)=
\frac{\sum_{y\in R_\delta(x)}\lambda^{d(x,y)}\,P(y)}
     {\sum_{y\in R_\delta(x)}\lambda^{d(x,y)}},
\]
where:
\begin{itemize}
  \item \(R_\delta(x)=\{\,y\mid d(x,y)\le\delta\}\) is the local neighborhood,
  \item \(d(\cdot,\cdot)\) is grid or graph distance,
  \item \(\lambda\in(0,1)\) controls spatial decay,
  \item \(P(y)\) is the composite potential at node \(y\).
\end{itemize}
This operation computes a weighted average of \(P\) over all nodes within radius \(\delta\), suppressing narrow, localized spikes (high‐frequency noise) while preserving broader, low‐frequency gradients.
The guard then selects
\[
n_{t+1}=\arg\min_{x\in\mathcal{C}(n_t)}K(x)
\]
as its next destination. The selection of the candidate node minimizing (K(x)) directs the guard towards regions with strong attractive signals (e.g., high negative information potential from the player) while balancing the influence of repulsive fields (e.g., high confidence in recently patrolled areas) and topological preferences, thereby achieving a decision that blends pursuit, coverage, and strategic navigation.

\subsection{Adaptive Weight Scheduling}
Weights are updated globally each frame:
\[
w_I(t)=
\begin{cases}
w_I^{\max}, & \text{if player detected},\\[4pt]
\max\bigl(w_I^{\text{base}},\,w_I(t\!-\!1)-\delta \Delta t\bigr),
 & \text{otherwise}.
\end{cases}
\]

Thus, $w_I(t)$ is maximized upon player detection; otherwise, it temporally decays towards $w_I^{\text{base}}$ by $\delta_w \Delta t$ each update. The remaining budget $1-w_I(t)$ is split between $w_C(t)$ and $w_N(t)$ according to their predefined base ratio, enabling smooth oscillation between \emph{alert} and \emph{patrol} modes without an explicit FSM.

\section{Experiments}

\subsection{Overview}
We evaluate our guard AI under two scenarios with distinct objectives. The \emph{Capture Experiment} assesses short‐term pursuit efficiency under heightened difficulty, while the \emph{Fixed‐Time Experiment} measures long‐term patrol stability in a standard setting. All tests combine two environment abstractions (Grid, Partition), five guard control modes (PotentialField, RandomWalk, FSM, Staleness, Cheat), two player policies (MaxMin, SafetyGraph), and five distinct levels.

\subsection{Scenario Details}
\label{sec:scenario_details}
We configure two distinct experiment scenarios with different difficulty levels and duration settings, designed to evaluate complementary aspects of guard AI performance.

\paragraph{Capture Experiment}
This scenario emphasizes short-term pursuit efficiency under heightened challenge. Players move at 120\% of their base speed, while guards' view range is reduced to 80\% of their base range, making the player more agile and harder to detect. Player health decreases by one per second for each guard whose attack range encompasses the player. Each trial in this experiment concludes either when the player's health reaches zero (i.e., captured) or after a maximum duration of 60 seconds. This setup measures how rapidly and reliably different guard AI strategies can detect and neutralize a challenging target under pressure.

\paragraph{Fixed‐Time Experiment}
In contrast, this scenario focuses on long-term patrol stability and sustained engagement in a standard gameplay setting. Players and guards operate at 100\% base speed and 100\% base view range. Each simulation trial runs for a fixed duration of exactly 60s, regardless of the player's health status or whether a capture occurs. The objective here is to assess whether guard strategies maintain effective, evenly distributed patrol patterns and consistent pursuit capabilities over an extended period.

\begin{table*}[htbp] 
\centering
\small
\resizebox{\textwidth}{!}{
\begin{tabular}{lcrrrccr}
\toprule
\textbf{Level Name} & \multicolumn{4}{c}{\textbf{NavMesh Properties}} & \multicolumn{3}{c}{\textbf{Grid Properties}} \\
\cmidrule(lr){2-5} \cmidrule(lr){6-8}
& Triangles & Min Area & Max Area & Avg Area & Dimensions & Walkable (0s) & Non-walkable (1s) \\
& (Count) & (world units²) & (world units²) & (world units²) & (Rows $\times$ Cols) & (Count (\%)) & (Count (\%)) \\
\midrule
Arkham & 140 & 0.1111 & 20.5000 & 2.8789 & 28 $\times$ 26 & 548 (75\%) & 180 (25\%) \\
Dishonored & 238 & 0.0833 & 15.3472 & 2.0090 & 26 $\times$ 39 & 726 (72\%) & 288 (28\%) \\
Miami & 178 & 0.0417 & 9.2083 & 1.1891 & 26 $\times$ 21 & 377 (69\%) & 169 (31\%) \\
Test (Maze) & 357 & 0.0025 & 6.9525 & 0.2220 & 20 $\times$ 20 & 206 (52\%) & 194 (48\%) \\ 
TLOU & 437 & 0.0017 & 2.5359 & 0.3008 & 14 $\times$ 30 & 255 (61\%) & 165 (39\%) \\
\bottomrule
\end{tabular}
}
\caption{Characteristics of the five test levels used in the experiments, detailing NavMesh properties (total triangles, min/max/avg triangle area) and Grid properties (dimensions, count and percentage of walkable (0) vs. non-walkable (1) cells).}
\label{tab:level_characteristics}
\end{table*}

\subsection{Guard Control Modes}
As the internal decision logic of commercial guards is closed-source, we approximate them with simplified baselines. We evaluate five guard-AI methods:
\begin{itemize}
    \item \textbf{Potential Field}: Our proposed method combines dynamic information, confidence, and connectivity fields into a kernel-filtered potential to guide guard movement. Guards continuously adapt their paths to pursue the player and efficiently patrol the area.
    \item \textbf{Random Walk}: Each guard randomly selects one reachable neighboring node or cell at each decision tick, disregarding previous states or player positions.
    \item \textbf{Finite State Machine (FSM)}: Guards switch explicitly between predefined states (Patrol, Investigate, Chase, Return; see Table~\ref{tab:experiment_parameters} for details). They patrol random paths until detecting a player, triggering investigation and pursuit behaviors based on set thresholds.
    \item \textbf{Staleness-based FSM}: Guards maintain a ``staleness'' value (time since last patrol) for each node. The FSM selects the next patrol location based on nodes with the highest staleness, ensuring frequent visits to neglected areas while still enabling standard Investigate–Chase transitions when encountering the player.
    \item \textbf{Cheat}: Guards directly follow the shortest possible path toward the player’s current location, serving as an idealized baseline for theoretical optimal performance.
\end{itemize}

\subsection{Player Policies}
We implement two player-control strategies——MaxMin, a simple yet effective greedy policy, and SafetyGraph, a longer-horizon planning approach—serving as reproducible proxies for human players to evaluate guard effectiveness:
\begin{itemize}
    \item \textbf{MaxMin Policy}: At each decision step, the player agent simulates moves several steps ahead, evaluating each potential path by calculating the minimum distance to any guard along that path. The agent chooses the move that maximizes this minimum distance, continuously aiming to stay as far as possible from all guards.
    
    \item \textbf{Safety Graph Policy}: We construct a ``safety graph'' by computing two sets of shortest-path distances: from each guard to every node (estimating guard arrival time \(T_g\)), and from the player to every node (player arrival time \(T_p\)). Each node’s safety is quantified by \(S=T_g - T_p\), indicating the time margin before potential capture. The player then plans paths using A* search, penalizing paths passing through nodes with lower safety margins, thus preferring routes with lower capture risk.
\end{itemize}

\subsection{Test Levels}
We test on five maps: a handcrafted Maze (``Test''), Arkham Asylum’s Medical Facility (``Arkham''), Dishonored\,2’s Grand Palace (``Dishonored''), Hitman\,2’s Miami (``Miami''), and a Seattle Bridge scene from The Last of Us\,2 (``TLOU''). Each layout is approximated by a binary 0/1 grid (for the Grid environment) or baked into a NavMesh triangulation (for the Partition environment). Key characteristics of these levels are summarized in Table~\ref{tab:level_characteristics}.

The selected levels present a diverse set of challenges for robust AI evaluation. Key factors include overall scale (from the compact ``TLOU'' grid to the larger ``Dishonored''), NavMesh granularity (e.g., ``Test (Maze)'' and ``TLOU'' feature dense NavMeshes with numerous small triangles, implying intricate paths, contrasting with ``Arkham'''s sparser, larger triangles), and spatial density (e.g., the ``Test (Maze)'' offers a near 50/50 split of walkable/non-walkable grid cells, indicating a dense, confined structure, while ``Arkham'' and ``Dishonored'' provide more open environments with ~72-75\% walkable area). This diversity in size, navigational complexity, and pathway constructiveness allows for a comprehensive assessment of how different guard AI, especially our Potential Field method, generalize and adapt their behaviors across varied spatial configurations.

\begin{figure*}[!t]
  \centering
  \subfloat[Grid Environment]{%
    \includegraphics[width=0.48\linewidth]{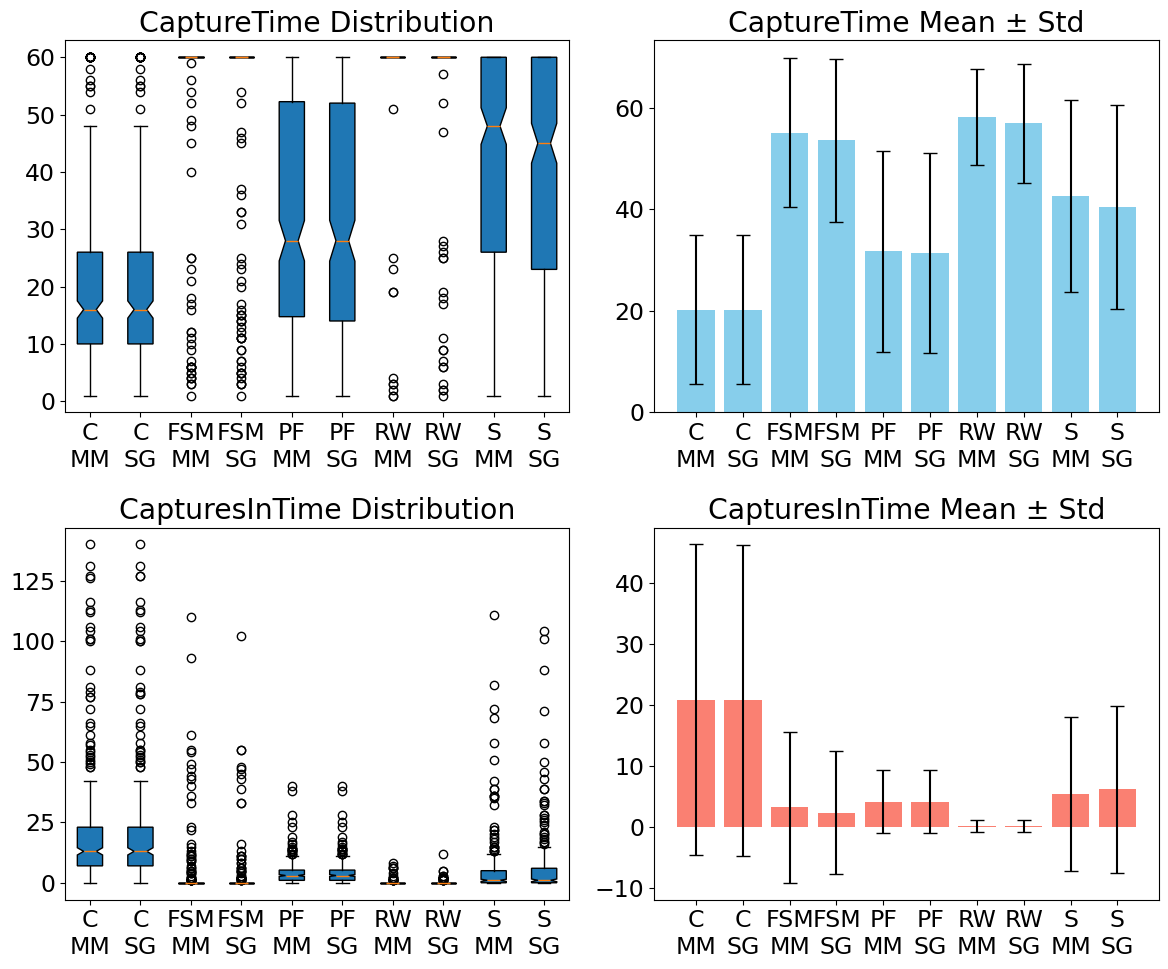}
    }
  \hfill
  \subfloat[Partition Environment]{%
    \includegraphics[width=0.48\linewidth]{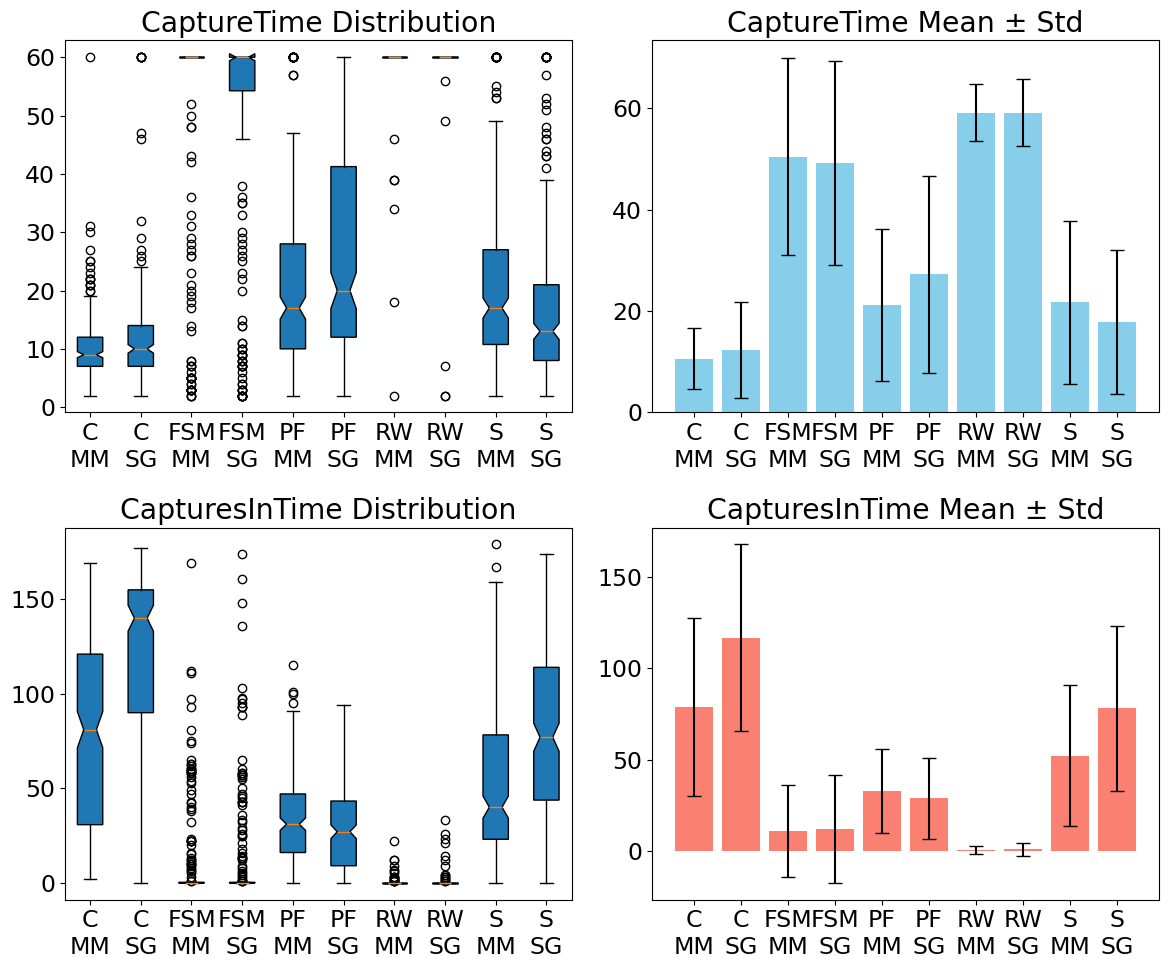}
    }
  \caption{Aggregated performance across all levels and player policies. Abbreviations: C = Cheat, MM = MaxMin, SG = SafetyGraph, PF = PotentialField, RW = RandomWalk, S = Staleness.}
  \label{fig:aggregate-results}
\end{figure*}

\subsection{Implementation Details and Parameters}
\label{sec:implementation_details}

Our experiments were conducted using a specific set of parameters for the proposed Potential Field method, as well as for the baseline FSM and Staleness-based FSM methods. Key parameters are summarized in Table~\ref{tab:experiment_parameters}. Player agents were initialized with 3 health points (HP). For each experimental repeat, the player was initialized at a grid cell or NavMesh node chosen to maximize its minimum shortest-path distance to any guard, with the additional constraint that the starting position was not within any guard's initial line of sight. This was intended to provide a fair and consistently challenging starting configuration. The simulation update interval for AI decisions was set to 1 second.

For each (environment, guard mode, player policy, level) combination we run $R=50$ repeats with independent random seeds. We record per-trial metrics—capture time or count, coverage rate, average guard–player distance, first detection time, backtracking count—and export Visit/Attack heatmaps as CSV for post hoc analysis.

\begin{table}[htbp]
\centering
\small
\begin{tabular}{llr}
\toprule
\textbf{Category} & \textbf{Parameter} & \textbf{Value} \\
\midrule
\multicolumn{3}{l}{\textit{Potential Field Parameters (Our Method)}} \\
& Information Init ($I_0$) & 5.0 \\
& Confidence Init ($C_0$) & 1.0 \\
& Information Decay ($\gamma_I$) & 0.9 \\
& Confidence Decay ($\gamma_C$) & 0.8 \\
& Field (Temporal) Decay ($\rho$) & 0.95 \\
& Kernel Lambda ($\lambda$) & 0.8 \\
& Kernel Neighborhood ($\delta$) & 3 \\
\midrule
\multicolumn{3}{l}{\textit{Dynamic Weight Parameters (Our Method)}} \\
& Basic Info Weight ($w_I^{\text{base}}$) & 0.4 \\
& Basic Conf Weight ($w_C^{\text{patrol}}$) & 0.5 \\
& Basic Conn Weight ($w_N^{\text{patrol}}$) & 0.1 \\
& Max Info Weight ($w_I^{\max}$) & 0.9 \\
& Weight Decay Rate ($\delta_w$) & 0.05 \\ 
\midrule
\multicolumn{3}{l}{\textit{FSM Parameters (Baseline)}} \\
& Info Thresh & 1 \\
& T Investigate & 3s \\ 
& T Lose Sight & 2s \\
& Chase Max Time & 10s \\
& Return Max Time & 5s \\
& Decision Interval & 1s \\
\midrule
\multicolumn{3}{l}{\textit{General Simulation}} \\
& Staleness Increment & 1 \\
& Player Health & 3 HP \\
& AI Update Interval & 1s \\
\bottomrule
\end{tabular}
\caption{Key experimental parameters for different AI components and simulation settings.}
\label{tab:experiment_parameters}
\end{table}

\section{Results}

We evaluate guard AI performance using several key metrics, each highlighting distinct aspects of effectiveness and behavior quality:
\begin{itemize}
    \item \textbf{CaptureRate} (Capture Exp.): Proportion of successful captures, measuring primary objective achievement.
    \item \textbf{CaptureTime} (Capture Exp.): Time until capture in successful trials; lower values indicate faster pursuit.
    \item \textbf{CapturesInTime} (Fixed-Time Exp.): Total player captures, reflecting sustained engagement capability.
    \item \textbf{CoverageRate}: Percentage of unique map areas visited by guards, indicating patrol thoroughness.
    \item \textbf{AvgGuard–PlayerDist}: Average closest distance between a guard and the player, reflecting pursuit closeness.
    \item \textbf{FirstDetectTime}: Time until the player is first detected, measuring guard vigilance and responsiveness.
    \item \textbf{BacktrackCount}: Frequency of guards immediately revisiting recent locations (e.g., within the last $N=5$ steps); lower values suggest more naturalistic and less redundant patrol paths.
\end{itemize}
Together, these metrics provide a multifaceted view of guard performance, from immediate threat response to sustained environmental control.

\begin{table*}[!t]
  \centering
  \scriptsize
  \resizebox{\textwidth}{!}{%
  \begin{tabular}{llccccccc}
    \toprule
    & & \multicolumn{2}{c}{\textbf{Capture Exp}} & \multicolumn{5}{c}{\textbf{Fixed‐Time Exp}} \\
    \cmidrule(lr){3-4}\cmidrule(lr){5-9}
    GuardMode & PlayerMode
      & CaptureRate
      & CaptureTime
      & CapturesInTime
      & CoverageRate
      & AvgGuard–PlayerDist
      & FirstDetectTime
      & BacktrackCount \\
    \midrule
    \multicolumn{9}{l}{\textbf{Grid Environment}} \\
    Cheat          & MaxMin       & 0.95 & 20.25\,$\pm$\,14.66 & 20.85\,$\pm$\,25.51 & 0.76\,$\pm$\,0.16 &  8.26\,$\pm$\,2.99 & 11.46\,$\pm$\,9.97 & 15.60\,$\pm$\,29.36 \\
    Cheat          & SafetyGraph  & 0.95 & 20.23\,$\pm$\,14.68 & 20.84\,$\pm$\,25.52 & 0.75\,$\pm$\,0.16 &  8.27\,$\pm$\,2.99 & 11.36\,$\pm$\,9.96 & 15.60\,$\pm$\,29.36 \\
    FSM            & MaxMin       & 0.12 & 55.10\,$\pm$\,14.68 &  3.23\,$\pm$\,12.37 & 0.61\,$\pm$\,0.15 & 17.80\,$\pm$\,5.86 & 47.46\,$\pm$\,22.71 & 53.34\,$\pm$\,14.19 \\
    FSM            & SafetyGraph  & 0.15 & 53.60\,$\pm$\,16.08 &  2.38\,$\pm$\,10.08 & 0.61\,$\pm$\,0.14 & 18.29\,$\pm$\,5.96 & 45.03\,$\pm$\,23.97 & 54.65\,$\pm$\,14.16 \\
    PotentialField & MaxMin       &  \textbf{0.80} & \textbf{31.70\,$\pm$\,19.72} &  4.18\,$\pm$\,5.24  & \textbf{0.94\,$\pm$\,0.10} & \textbf{15.17\,$\pm$\,5.26} & \textbf{15.16\,$\pm$\,15.07} &  \textbf{2.94\,$\pm$\,3.06}  \\
    PotentialField & SafetyGraph  &  \textbf{0.80} & \textbf{31.39\,$\pm$\,19.78} &  4.19\,$\pm$\,5.22  & \textbf{0.94\,$\pm$\,0.10} & \textbf{15.18\,$\pm$\,5.25} & \textbf{15.10\,$\pm$\,15.06} &  \textbf{2.94\,$\pm$\,3.06}  \\
    RandomWalk     & MaxMin       & 0.04 & 58.18\,$\pm$\,9.48  &  0.20\,$\pm$\,0.96  & 0.61\,$\pm$\,0.14 & 17.90\,$\pm$\,5.80 & 46.18\,$\pm$\,23.18 & 56.48\,$\pm$\,12.96 \\
    RandomWalk     & SafetyGraph  & 0.08 & 56.92\,$\pm$\,11.72 &  0.19\,$\pm$\,0.94  & 0.62\,$\pm$\,0.15 & 18.23\,$\pm$\,5.96 & 45.45\,$\pm$\,23.19 & 55.45\,$\pm$\,13.15 \\
    Staleness      & MaxMin       & 0.56 & 42.65\,$\pm$\,18.90 &  \textbf{5.45\,$\pm$\,12.69} & 0.94\,$\pm$\,0.08 & 15.74\,$\pm$\,4.30 & 17.29\,$\pm$\,17.02 &  9.14\,$\pm$\,13.67 \\
    Staleness      & SafetyGraph  & 0.62 & 40.51\,$\pm$\,20.08 &  \textbf{6.18\,$\pm$\,13.67} & 0.94\,$\pm$\,0.08 & 15.72\,$\pm$\,4.53 & 15.35\,$\pm$\,16.26 &  9.48\,$\pm$\,13.49 \\
    \midrule
    \multicolumn{9}{l}{\textbf{Partition Environment}} \\
    Cheat          & MaxMin       & 1.00 & 10.61\,$\pm$\,6.00  & 78.81\,$\pm$\,48.61 & 1.21\,$\pm$\,0.23 &  3.30\,$\pm$\,1.09 &  4.18\,$\pm$\,3.13 & 44.20\,$\pm$\,48.64 \\
    Cheat          & SafetyGraph  & 0.98 & 12.33\,$\pm$\,9.53  &116.76\,$\pm$\,51.15 & 1.18\,$\pm$\,0.20 &  3.00\,$\pm$\,1.43 &  4.88\,$\pm$\,4.00 & 97.29\,$\pm$\,49.90 \\
    FSM            & MaxMin       & 0.22 & 50.45\,$\pm$\,19.46 & 10.77\,$\pm$\,25.12 & 0.99\,$\pm$\,0.27 & 14.85\,$\pm$\,5.05 & 44.86\,$\pm$\,23.69 & 64.22\,$\pm$\,29.66 \\
    FSM            & SafetyGraph  & 0.27 & 49.17\,$\pm$\,20.13 & 12.07\,$\pm$\,29.66 & 0.98\,$\pm$\,0.26 & 14.31\,$\pm$\,5.16 & 46.39\,$\pm$\,23.49 & 61.72\,$\pm$\,30.65 \\
    PotentialField & MaxMin       &  \textbf{0.93} &  \textbf{21.10\,$\pm$\,14.99} & 32.79\,$\pm$\,22.97 &  \textbf{1.72\,$\pm$\,0.23} &  9.02\,$\pm$\,4.76 &  9.95\,$\pm$\,15.04 &  \textbf{16.69\,$\pm$\,13.89} \\
    PotentialField & SafetyGraph  & 0.82 & 27.21\,$\pm$\,19.36 & 28.78\,$\pm$\,22.19 &  \textbf{1.73\,$\pm$\,0.21} &  9.62\,$\pm$\,3.90 & \textbf{9.43\,$\pm$\,13.58} &  \textbf{19.43\,$\pm$\,15.11} \\
    RandomWalk     & MaxMin       & 0.03 & 59.17\,$\pm$\,5.54  &  0.42\,$\pm$\,2.10  & 0.99\,$\pm$\,0.26 & 15.12\,$\pm$\,4.75 & 45.86\,$\pm$\,23.44 & 76.60\,$\pm$\,15.07 \\
    RandomWalk     & SafetyGraph  & 0.02 & 59.16\,$\pm$\,6.59  &  0.72\,$\pm$\,3.76  & 0.99\,$\pm$\,0.24 & 14.47\,$\pm$\,4.77 & 46.20\,$\pm$\,22.96 & 75.54\,$\pm$\,14.17 \\
    Staleness      & MaxMin       & 0.91 & 21.72\,$\pm$\,16.15 & \textbf{52.17\,$\pm$\,38.85} & 1.72\,$\pm$\,0.22 &  \textbf{8.58\,$\pm$\,3.78} & \textbf{9.35\,$\pm$\,9.26} &165.86\,$\pm$\,25.95\\
    Staleness      & SafetyGraph  &  \textbf{0.95} &  \textbf{17.90\,$\pm$\,14.24} &  \textbf{78.08\,$\pm$\,45.30} & 1.61\,$\pm$\,0.23 &  \textbf{8.14\,$\pm$\,4.37} &  9.77\,$\pm$\,9.55 &165.86\,$\pm$\,25.95 \\
    \bottomrule
  \end{tabular}}
  \caption{Summary of CaptureRate/CaptureTime and all Fixed‐Time metrics
    for both Grid and Partition environments.  All values are
    mean\,$\pm$\,std (two decimals).}
  \label{tab:fixed-based-summary}
\end{table*}

\subsection{Aggregated Performance Analysis}

Figure~\ref{fig:aggregate-results} and Table~\ref{tab:fixed-based-summary} summarize results over all five levels and both player policies, for the Grid and Partition variants.  We compare our Potential Field method against the idealized Cheat baseline and three realistic baselines: FSM, RandomWalk, and Staleness.

\paragraph{Capture Experiment}  
In the Grid environment (Fig.~\ref{fig:aggregate-results}(a), top), PF achieves a \emph{Capture Rate} of 0.80 with mean \emph{Capture Time} 31.7\,$\pm$\,19.7s, significantly faster than FSM (55.1\,$\pm$\,14.7s, rate 0.12), RandomWalk (58.2\,$\pm$\,9.5s, rate 0.04) and Staleness (42.7\,$\pm$\,18.9s, rate 0.56).  Although Cheat remains the lower bound (20.3\,$\pm$\,14.7s, rate 0.95), PF closes over half that gap compared to classical baselines, confirming that the combination of information, confidence and connectivity fields yields much more effective pursuit under pressure.

In the Partition environment (Fig.~\ref{fig:aggregate-results}(b), top), PF captures at 21.1\,$\pm$\,15.0s with rate 0.93, again outperforming FSM (50.5\,$\pm$\,19.5s, rate 0.22), RandomWalk (59.2\,$\pm$\,5.5s, rate 0.03), and even Staleness (21.7\,$\pm$\,16.2s, rate 0.91).  The decaying confidence map prevents guards from clustering prematurely and fosters broad flanking, offering both speed and robustness in complex navmesh layouts.

\paragraph{Fixed‐Time Experiment}  
Over 60s in the Grid variant (Fig.~\ref{fig:aggregate-results}(a), bottom), PF records 4.18\,$\pm$\,5.24 captures, coverage 0.945\,$\pm$\,0.097, average guard–player distance 15.17\,$\pm$\,5.26m, and first detection at 15.16\,$\pm$\,15.07s.  Staleness FSM slightly exceeds PF in total captures (5.45\,$\pm$\,12.69) but at the cost of heavy backtracking (9.14\,$\pm$\,13.67 vs.\ PF’s 2.94\,$\pm$\,3.06), indicating less efficient routing and more repetitive motion.

In the Partition long‐run (Fig.~\ref{fig:aggregate-results}(b), bottom), PF yields 32.79\,$\pm$\,22.97 captures, coverage 1.72\,$\pm$\,0.23, distance 9.02\,$\pm$\,4.76m, and detection at 9.95\,$\pm$\,15.04s.  Staleness FSM captures more (52.17\,$\pm$\,38.85) and covers slightly more area (1.72\,$\pm$\,0.22) but incurs extremely high backtracking (165.86\,$\pm$\,25.95), reflecting an unnatural ``vacuum‐cleaner'' revisit pattern.  PF strikes a balance: distributed patrols that still converge efficiently on the player without excessive zig‐zagging.

\paragraph{Analysis of PF’s Behavior}  
The confidence decay in PF encourages guards to spread out and maintain coverage, while the information field quickly refocuses effort once the player is detected—this leads to high capture efficiency in the short‐run and plausible patrols over time. This emergent ``encircle-and-squeeze'' pattern yields fast first captures yet avoids over-concentration, explaining why PF trails Staleness on the \emph{CapturesInTime} count but still posts better coverage and fewer backtracks. As Table~\ref{tab:fixed-based-summary} shows, PF also slightly outperforms Staleness in \emph{AvgGuard–Player Distance} (Grid: 15.17 vs.\ 15.74m; Partition: 9.02 vs.\ 8.58m) and \emph{First Detect Time} (Grid: 15.16 vs.\ 17.29s; Partition: 9.95 vs.\ 9.35s) while maintaining the lowest \emph{Backtrack Count}. More aggressive tuning (e.g.\ boosting information weight upon sighting and slowing its decay) could further tighten pursuit, but naturalistic guard motion is favored to avoid the extreme clustering of Cheat or the repetitive loops of Staleness FSM.  Overall, PF significantly outperforms classical baselines and approaches the theoretical limit set by Cheat, across both environments and both experimental conditions. 

In the Partition environment, uneven NavMesh triangles impose extra replanning overhead, slightly slowing capture times; nonetheless, PF still performs competitively even on Unity’s default mesh, confirming its robustness. By contrast, Staleness patrols are largely unaffected by mesh quality. We center our discussion on the MaxMin policy—its straightforward ``maximize distance'' rule leads to stable, interpretable behavior—whereas SafetyGraph trials occasionally stall in dense triangulations (i.e.\ in the partition environment, the abundance of near-equivalent shortest paths forces repeated tie-breaking, and our Euclidean heuristic provides little guidance, leading to prolonged A* expansions), as shown by its higher variability in key metrics in Table \ref{tab:fixed-based-summary}, so we report but do not specifically analyze those results here.

\begin{figure*}[!t]
  \centering
  \subfloat[Full Miami layout (used for both Grid and Partition).]{
    \includegraphics[width=0.335\linewidth]{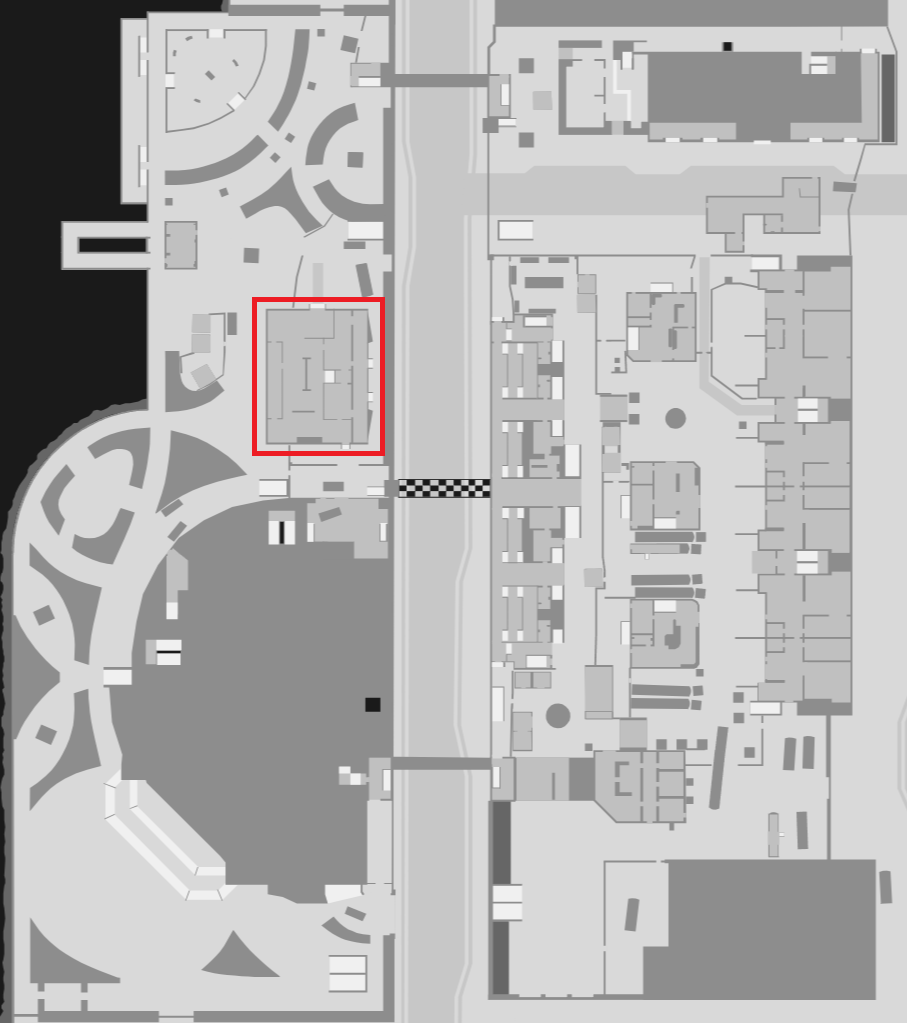}
    \label{fig:miami-big}
  }\hfill
  \subfloat[$26\times21$ occupancy grid extract.]{%
    \includegraphics[width=0.285\linewidth]{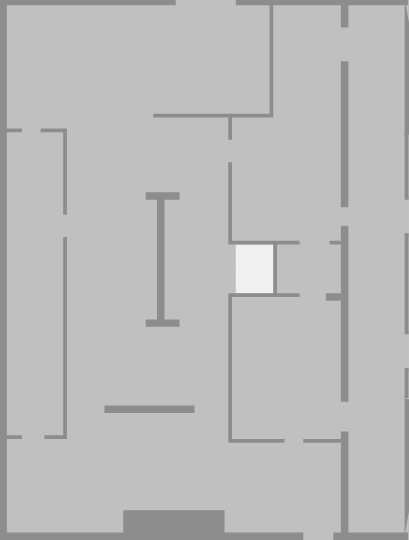}
    \label{fig:miami-small}
  }\hfill
  \subfloat[NavMesh partition graph (blue=centroids, red=edges).]{
    \includegraphics[width=0.305\linewidth]{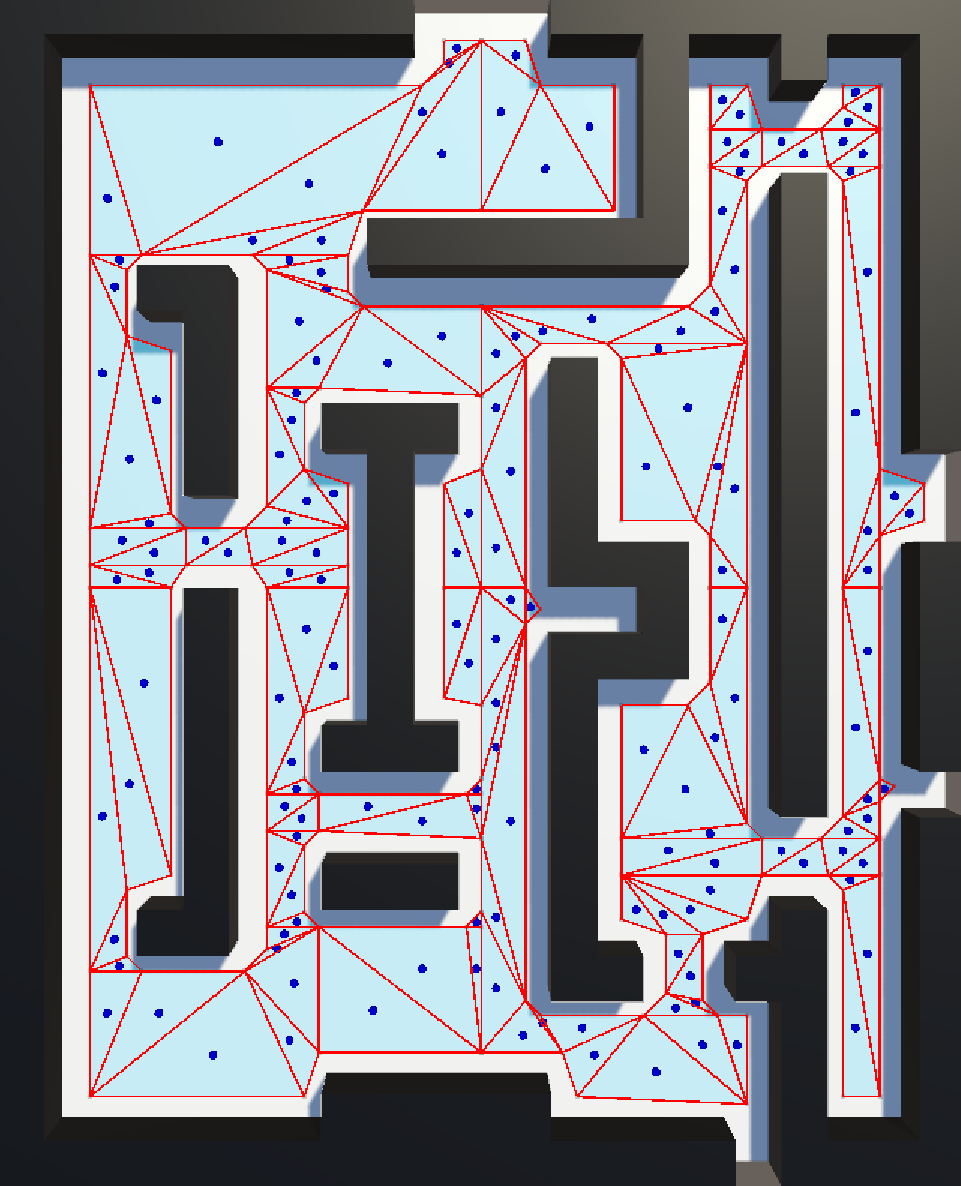}
    \label{fig:miami-par}
  }
  \caption{\textbf{Miami map assets} (a) The full layout. (b) Discretized grid for the Grid experiments. (c) NavMesh‐derived graph for the Partition experiments.}
  \label{fig:miami-assets}
\end{figure*}

\subsection{Heatmap Case Study: Miami Level}

\begin{figure*}[!t]
  \centering
  \subfloat[Visit Heatmaps — Miami]{
    \includegraphics[width=0.48\linewidth]{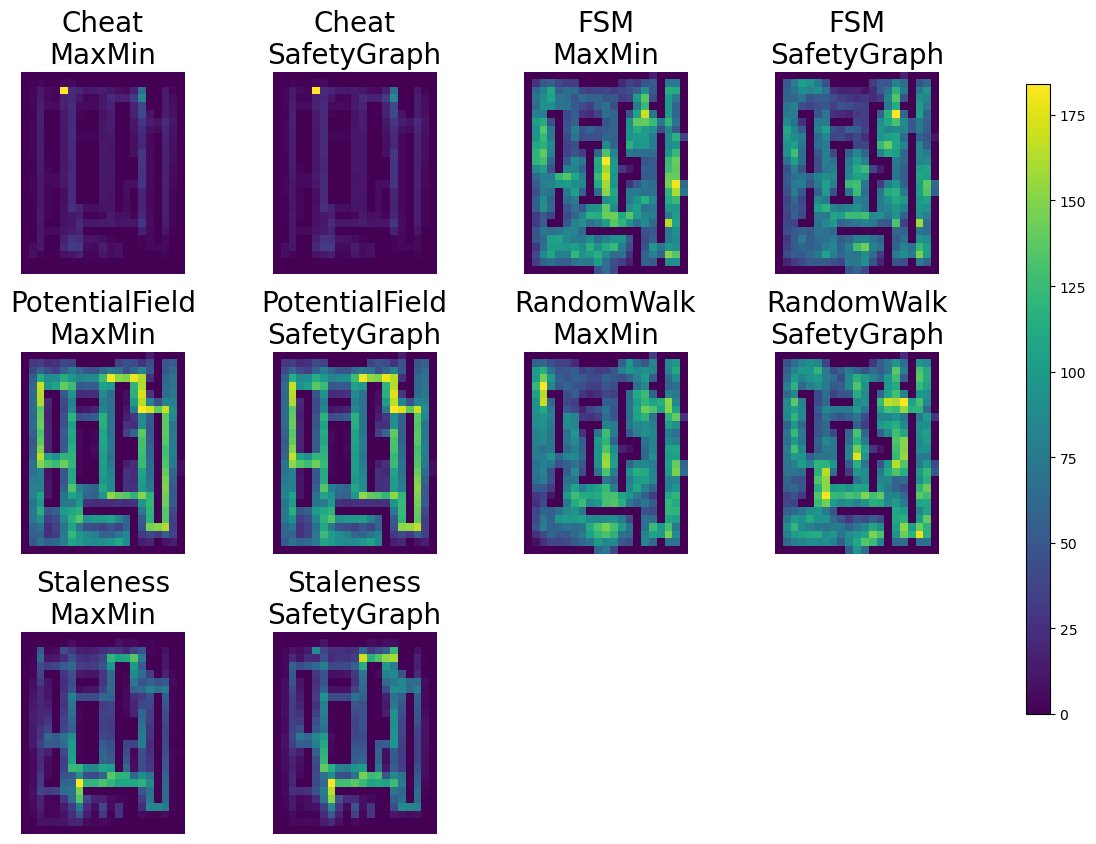}%
    \label{fig:heatmap-visit}
  }\hfill
  \subfloat[Attack Heatmaps — Miami]{
    \includegraphics[width=0.48\linewidth]{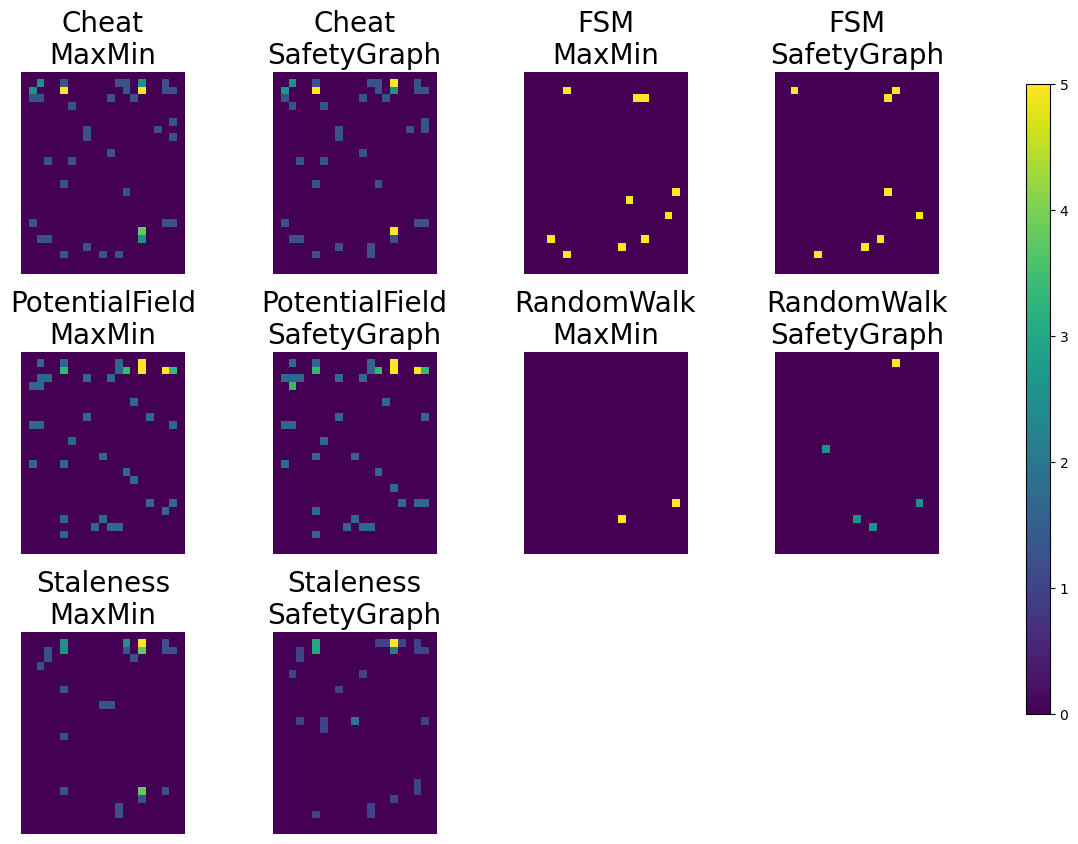}%
    \label{fig:heatmap-attack}
  }
  \caption{\textbf{Visit and Attack Heatmaps} on the Miami level (Grid environment).  Brighter cells indicate higher visit or capture counts across all guard–player policy combinations.}
  \label{fig:heatmap-miami}
\end{figure*}

To ground our aggregate results in spatial context, we first present the Miami level geometry used in both experiments (Fig.~\ref{fig:miami-assets}), and then overlay guard \emph{visit} and \emph{attack} heatmaps (Fig.~\ref{fig:heatmap-miami}). Figure~\ref{fig:miami-big} shows the entire Miami level. In the Grid variant (Fig.~\ref{fig:miami-small}), walkable areas are rasterized into a $26\times21$ cell grid; in the Partition variant (Fig.~\ref{fig:miami-par}), we triangulate that same area and use triangle centroids as nodes. By keeping geometry identical, we isolate the effect of our composite‐field AI on both substrates.

\paragraph{Visit Patterns}  
As shown in Fig.~\ref{fig:heatmap-visit}, the \emph{Cheat} and \emph{Staleness} baselines concentrate nearly all visits along a few major corridors, neglecting side passages.  \emph{FSM} and \emph{RandomWalk} improve but still waste effort in dead‐end branches. In contrast, our PF variants (MaxMin and SafetyGraph) have a more uniform coverage: main routes receive strong attention—mirroring Cheat’s axes—while secondary corridors and junctions are also patrolled regularly.  This validates PF’s ability to avoid both the tunnel‐vision of \emph{Cheat}/\emph{Staleness} and the aimless wandering of simpler baselines.

\paragraph{Attack Locations}  
Figure~\ref{fig:heatmap-attack} reveals that PF’s capture spots align with \emph{Cheat} hotspots—validating its aggressive pursuit—yet PF also secures additional flanking captures in side passages that \emph{Cheat} rarely uses.  \emph{FSM} and \emph{RandomWalk} produce sparse, rare attacks, while \emph{Staleness} seldom captures outside core corridors.  Overall, PF not only tracks the player reliably but also leverages distributed coverage to corral the player from multiple approaches, combining strategic focus and spatial control.

\subsection{Extensible Stealth‐Mechanism Simulation}

To show the adaptability of our composite potential‐field framework, we manually implemented and briefly validated seven common stealth‐game mechanics.  Each uses simple extensions to our Information and Confidence fields; no large‐scale automated tests were performed, only hand‐guided simulations to confirm plausibility.

\paragraph{Footstep Noise}  
We maintain a temporary Information field $I_{\text{temp}}$ in addition to the global $I$.  Whenever the player sprints, we inject at location $x_{s}$:
\[
  I_{\text{temp}}(y)\;\mathrel{+}=A_{n}\,\alpha^{\,d(x_{s},y)},\quad
  A_{n}=3.0,\;\alpha=0.6,\;d\le5.
\]
Each second $I_{\text{temp}}(y)\leftarrow0.8\cdot I_{\text{temp}}(y)$.  In our manual simulation, guards briefly veer toward the sprint path and then resume patrol as $I_{\text{temp}}$ decays below 0.1.

\paragraph{Thrown Decoy}  
On throw at position $x_{t}$, we spawn a one‐off impulse in $I_{\text{temp}}$ identical to Footstep Noise, then let it decay.  We gave the player a 10\,s cooldown and allowed placement within a Manhattan radius of 4 cells (avoiding walls).  Guards reliably investigate the landing spot before refocusing on normal patrol.

\paragraph{Static Decoy}  
We precompute a persistent ``static'' Information map $I_{\text{static}}$ by seeding
\[
  I_{\text{static}}(y)=A_{d}\,\beta^{\,d(x_{d},y)},\quad A_{d}=5.0,\;\beta=0.8,
\]
over the entire grid or graph at game start.  Once the player interacts at $x_{d}$, we add $I_{\text{static}}$ into the global $I$ (with full‐map BFS), causing guards to converge.  After the first guard arrives and ``disarms'' the decoy for 3\,s, we remove $I_{\text{static}}$ and observe guards return to baseline patrol.

\paragraph{Corpse Discovery}  
When the player ``kills'' a guard at $x_{c}$, we create another static map $I_{\text{corpse}}$ with
\[
  I_{\text{corpse}}(y)=A_{c}\,\gamma^{\,t}\,\delta^{\,d(x_{c},y)},
  \quad A_{c}=4.0,\;\gamma=0.95\!/{\rm s},\;\delta=0.7.
\]
Only after a guard reaches $x_{c}$ and spends 10\,s ``inspecting'' do we remove $I_{\text{corpse}}$.  In our test, this triggered realistic crowding around the body before gradual dispersal.

\paragraph{Lighting Effects}  
In designated ``dark'' cells we override the global Confidence‐decay $\lambda_{C}=0.95$ by a faster rate $\lambda_{C}^{\prime}=0.80$.  That is,
\[
  C(y)\leftarrow
  \begin{cases}
    0.80\,C(y), & y\in\text{dark}\\
    0.95\,C(y), & \text{otherwise.}
  \end{cases}
\]
Manual trials showed guards re‐visit shadowed areas more frequently, simulating cautious search under low light.

\paragraph{Weather Effects}  
When ``raining,'' we halve all footstep amplitudes ($A_{n}\!=\!1.5$) and increase their decay to $\alpha=0.7$.  Guards in our simulation detected only very close footsteps, modeling sound dampening by rain.

\paragraph{Concealment (Grass, Smoke)}  
Cells tagged as ``concealing'' apply both Lighting‐style confidence decay ($\lambda_{C}^{\prime}=0.80$) and reduce each guard’s view range $r$ to $0.7r$.  We observed that guards paused at smoke boundaries, hesitated longer before entering tall grass zones, and exhibited believable uncertainty.

\vspace{0.5ex}
\noindent
Because these extensions merely adjust field‐update parameters and injection rules, they slot seamlessly into our existing AI loop.  Although we only performed hand‐guided playtests to verify each behavior, the framework clearly supports rapid prototyping of diverse stealth mechanisms without specific scripting or new state machines.

\section*{Conclusion}
In this paper, we introduced a novel, training-free, and explainable framework for generic guard AI in stealth games, centered around the concept of Composite Potential Fields. Our approach integrates three interpretable maps---Information, Confidence, and Connectivity---into a unified, kernel-filtered decision criterion, enabling guards to dynamically balance broad patrol coverage with responsive pursuit of the player. The parametric, designer-driven nature of our system allows for intuitive tuning and adaptation across different environmental abstractions, namely occupancy grids and NavMesh partitions, without the need for retraining.

Our experiments, conducted across five diverse game levels, two player policies, and compared against four distinct baseline guard behaviors, indicate the efficacy of our proposed Potential Field method. The results confirm that PF significantly outperforms classical FSM and Staleness-based approaches in terms of capture efficiency, particularly under high-pressure scenarios. Furthermore, our method achieves superior patrol naturalness, evidenced by higher coverage rates and substantially lower backtracking counts, indicating more plausible and less repetitive guard movements. The heatmap analyses provided qualitative validation of PF's ability to achieve both focused pursuit and comprehensive environmental coverage.

Beyond core patrol and pursuit, we proved the adaptability of our framework by extending it to simulate a variety of common stealth game mechanics, such as footstep noise, decoys, corpse discovery, and environmental effects like lighting and concealment. These extensions were achieved through simple parameter adjustments and additions to the existing field update rules, highlighting the framework's capacity for adaptive prototyping of dynamic guard behaviors.

\paragraph{Limitations} Despite these contributions, our work has limitations. 
The current parameter tuning for the adaptive weights and field dynamics, while effective, was performed manually and could potentially be optimized further, perhaps through a designer-assisted search or limited learning process for specific game styles. While the extensibility demonstrations were plausible, they were validated through hand-guided simulations rather than large-scale quantitative testing, and we have not evaluated how they impact human players in actual gameplay. Additionally, the current model primarily focuses on individual guard decision-making; explicit coordination strategies for multiple guards, beyond emergent behaviors from shared field information, remain an area for deeper exploration. Our Partition implementation also relies on Unity’s default NavMesh triangulation—easy to reproduce but vulnerable to uneven mesh quality.

\paragraph{Future works} Future work will focus on several key directions. 
First, we plan to investigate methods for semi-automated tuning of the parameters to better match specific designer goals or desired difficulty curves. Second, we aim to conduct formal user studies with human players to assess perceived naturalness, challenge, and immersion compared to existing AI techniques. Third, exploring better multi-guard coordination strategies built upon the composite potential field foundation is a promising avenue. This could involve guards sharing information more explicitly or taking on specialized roles within the composite field framework. Finally, we intend to apply and evaluate our framework in more complex, large-scale 3D environments with richer stealth mechanics and to evaluate alternative partition strategies—such as level skeletons or optimized mesh segmentation—to further enhance field fidelity and consistency.

We believe our Composite Potential Field framework offers a valuable contribution towards creating more intelligent, believable, and adaptable AI agents in stealth games, providing designers with tunable tools to enhance player immersion and strategic engagement. 

\bibliography{aaai25}

\begin{thebibliography}{20}
\providecommand{\natexlab}[1]{#1}

\bibitem[{Al~Enezi and Verbrugge(2020)}]{al2020dynamic}
Al~Enezi, W.; and Verbrugge, C. 2020.
\newblock Dynamic guard patrol in stealth games.
\newblock In \emph{Proceedings of the AAAI Conference on Artificial Intelligence and Interactive Digital Entertainment}, volume~16, 160--166.

\bibitem[{Al~Enezi and Verbrugge(2021)}]{9619054}
Al~Enezi, W.; and Verbrugge, C. 2021.
\newblock Skeleton-based multi-agent opponent search.
\newblock In \emph{2021 IEEE Conference on Games (CoG)}, 1--8.

\bibitem[{Al~Enezi and Verbrugge(2023{\natexlab{a}})}]{al2023evaluating}
Al~Enezi, W.; and Verbrugge, C. 2023{\natexlab{a}}.
\newblock Evaluating player experience in stealth games: Dynamic guard patrol behavior study.
\newblock In \emph{Proceedings of the AAAI Conference on Artificial Intelligence and Interactive Digital Entertainment}, volume~19, 175--184.

\bibitem[{Al~Enezi and Verbrugge(2023{\natexlab{b}})}]{al2023investigating}
Al~Enezi, W.; and Verbrugge, C. 2023{\natexlab{b}}.
\newblock Investigating the influence of behaviors and dialogs on player enjoyment in stealth games.
\newblock In \emph{Proceedings of the AAAI Conference on Artificial Intelligence and Interactive Digital Entertainment}, volume~19, 166--174.

\bibitem[{Alarabi et~al.(2024)Alarabi, Lei, Santora, Luo, and Sellers}]{alarabi2024multi}
Alarabi, S.; Lei, T.; Santora, M.; Luo, C.; and Sellers, T. 2024.
\newblock Multi-robot path planning using potential field-based simulated annealing approach.
\newblock In \emph{Unmanned Systems Technology XXVI}, volume 13055, 102--117. SPIE.

\bibitem[{Cern{\'a}k and Lianoudakis(2021)}]{cernak2021tower}
Cern{\'a}k, M.; and Lianoudakis, O. 2021.
\newblock \emph{The Tower: Design of Distinct AI Personalities in Stealth-Based Games Modeling personalities with believable and consistent behaviour for NPCs in games using stealth gameplay style}.
\newblock Master's thesis, Uppsala University.

\bibitem[{Chia(2022)}]{chia2022artist}
Chia, A. 2022.
\newblock The artist and the automaton in digital game production.
\newblock \emph{Convergence}, 28(2): 389--412.

\bibitem[{Danielsiek et~al.(2008)Danielsiek, Stuer, Thom, Beume, Naujoks, and Preuss}]{danielsiek2008intelligent}
Danielsiek, H.; Stuer, R.; Thom, A.; Beume, N.; Naujoks, B.; and Preuss, M. 2008.
\newblock Intelligent moving of groups in real-time strategy games.
\newblock In \emph{2008 IEEE Symposium On Computational Intelligence and Games}, 71--78. IEEE.

\bibitem[{Hagelb{\"a}ck and Johansson(2008)}]{hagelback2008using}
Hagelb{\"a}ck, J.; and Johansson, S.~J. 2008.
\newblock Using multi-agent potential fields in real-time strategy games.
\newblock In \emph{Seventh International Conference on Autonomous Agents and Multi-agent Systems (AAMAS), 12-16, 2008, Estoril}, 631--638.

\bibitem[{Helbing and Molnar(1995)}]{helbing1995social}
Helbing, D.; and Molnar, P. 1995.
\newblock Social force model for pedestrian dynamics.
\newblock \emph{Physical review E}, 51(5): 4282.

\bibitem[{Hladky and Bulitko(2008)}]{hladky2008evaluation}
Hladky, S.; and Bulitko, V. 2008.
\newblock An evaluation of models for predicting opponent positions in first-person shooter video games.
\newblock In \emph{2008 IEEE Symposium On Computational Intelligence and Games}, 39--46. IEEE.

\bibitem[{Isla(2005)}]{isla2005handling}
Isla, D. 2005.
\newblock {Handling Complexity in the Halo 2 AI System}.
\newblock Presented at the Game Developers Conference.
\newblock GDC Talk.

\bibitem[{Isla(2006)}]{isla2006probabilistic}
Isla, D. 2006.
\newblock Probabilistic target tracking and search using occupancy maps.
\newblock \emph{AI Game Programming Wisdom}, 3: 379--388.

\bibitem[{Isla(2013)}]{isla2013third}
Isla, D. 2013.
\newblock Third Eye Crime: Building a stealth game around occupancy maps.
\newblock In \emph{Proceedings of the AAAI Conference on Artificial Intelligence and Interactive Digital Entertainment}, volume~9, 206--206.

\bibitem[{Khatib(1986)}]{khatib1986real}
Khatib, O. 1986.
\newblock Real-time obstacle avoidance for manipulators and mobile robots.
\newblock \emph{The international journal of robotics research}, 5(1): 90--98.

\bibitem[{Li et~al.(2024)Li, Jiang, Liu, Zhang, Xu, and Liu}]{li2024generative}
Li, S.; Jiang, H.; Liu, Y.; Zhang, J.; Xu, X.; and Liu, D. 2024.
\newblock Generative subgoal oriented multi-agent reinforcement learning through potential field.
\newblock \emph{Neural Networks}, 179: 106552.

\bibitem[{Samodro, Puriyanto, and Caesarendra(2023)}]{samodro2023artificial}
Samodro, M. M.~J.; Puriyanto, R.~D.; and Caesarendra, W. 2023.
\newblock Artificial potential field path planning algorithm in differential drive mobile robot platform for dynamic environment.
\newblock \emph{International Journal of Robotics and Control Systems}, 3(2): 161--170.

\bibitem[{Sánchez-Ruiz and Miranda(2017)}]{SANCHEZRUIZ201729}
Sánchez-Ruiz, A.~A.; and Miranda, M. 2017.
\newblock A machine learning approach to predict the winner in StarCraft based on influence maps.
\newblock \emph{Entertainment Computing}, 19: 29--41.

\bibitem[{Uriarte and Ontan{\'o}n(2012)}]{uriarte2012kiting}
Uriarte, A.; and Ontan{\'o}n, S. 2012.
\newblock Kiting in RTS games using influence maps.
\newblock In \emph{Proceedings of the AAAI conference on artificial intelligence and interactive digital entertainment}, volume~8, 31--36.

\bibitem[{Xu, Tremblay, and Verbrugge(2014)}]{xu2014generative}
Xu, Q.; Tremblay, J.; and Verbrugge, C. 2014.
\newblock Generative methods for guard and camera placement in stealth games.
\newblock In \emph{Proceedings of the AAAI Conference on Artificial Intelligence and Interactive Digital Entertainment}, volume~10, 87--93.

\end{thebibliography}

\end{document}